\documentclass[letterpaper, 10 pt, conference]{ieeeconf}  % Comment this line out if you need a4paper

\IEEEoverridecommandlockouts                              % This command is only needed if 
\usepackage{cite} 
\usepackage{xcolor}
\usepackage{amsfonts}
\usepackage{amsmath}
\usepackage{booktabs}
\usepackage{multirow}
\usepackage{graphicx}
\usepackage{etoolbox}
\usepackage{capt-of}
\usepackage{hyperref}

\title{\LARGE \bf
ReactEMG Stroke: Healthy-to-Stroke Few-shot Adaptation \\ for sEMG-Based Intent Detection
}

\author{
Runsheng Wang$^{1}$, Katelyn Lee$^{1}$, Xinyue Zhu$^{2}$, Lauren Winterbottom$^{3}$,\\
Dawn M. Nilsen$^{3,*}$, Joel Stein$^{3,*}$, and Matei Ciocarlie$^{1,*}$%
\thanks{*Co-Principal Investigators.}%
\thanks{$^{1}$Department of Mechanical Engineering, Columbia University in the City of New York, NY, USA
{\tt\small \{runsheng.w, katelyn.lee, matei.ciocarlie\}@columbia.edu}}%
\thanks{$^{2}$Department of Computer Science, Columbia University in the City of New York, NY, USA
{\tt\small \{xz3013\}@columbia.edu}}%
\thanks{$^{3}$Department of Rehabilitation and Regenerative Medicine, Columbia University Irving Medical Center,
New York, NY 10032, USA {\tt\small \{lbw2136, dmn12, js1165\}@cumc.columbia.edu}}%
\thanks{This work was supported in part by an Amazon Research Award, by the National Science Foundation Graduate Research Fellowship Program under grant No. DGE-2437839, and by the Columbia Data Science Institute Seed Program.}%
}

\newcommand{\teaserfig}{%
  \begin{center}
    \includegraphics[width=\textwidth]{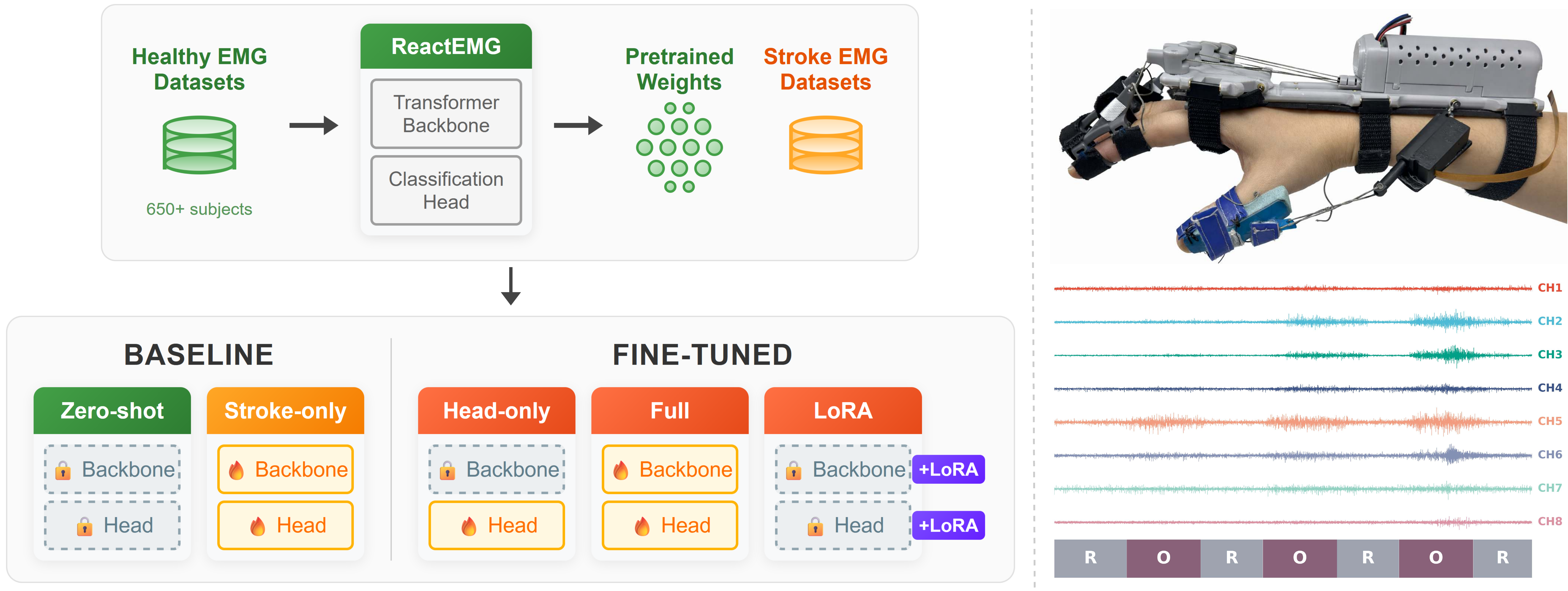}\par
    \captionof{figure}{We introduce \textbf{ReactEMG Stroke}, a model for predicting users' intent for the purpose of controlling a wearable hand orthosis. We pretrain our model on large-scale able-bodied sEMG data and adapt it to each stroke participant via fine-tuning on a small amount of labeled stroke data. We compare two baselines—healthy zero-shot transfer and stroke-only training from scratch—with three adaptation strategies: head-only tuning, full end-to-end fine-tuning, and parameter-efficient LoRA (locks/flames denote frozen/trainable parameters). Right: MyHand orthosis setup and example 8-channel stroke sEMG data during a cue sequence where \texttt{R} indicates intent to relax the hand and \texttt{O} indicates attempted hand opening.}
    \label{fig:intro}
  \end{center}
  \vspace{0.5em}
}

\makeatletter
\apptocmd{\@maketitle}{\teaserfig}{}{}
\makeatother

\begin{document}

\maketitle
\thispagestyle{empty}
\pagestyle{empty}

%%%%%%%%%%%%%%%%%%%%%%%%%%%%%%%%%%%%%%%%%%%%%%%%%%%%%%%%%%%%%%%%%%%%%%%%%%%%%%%%
%%%%%%%%%%%%%%%%%%%%%%%%%%%%%%%%%%%%%%%%%%%%%%%%%%%%%%%%%%%%%%%%%%%%%%%%%%%%%%%%
%%%%%%%%%%%%%%%%%%%%%%%%%%%%%%%%%%%%%%%%%%%%%%%%%%%%%%%%%%%%%%%%%%%%%%%%%%%%%%%%

\begin{abstract}
Surface electromyography (sEMG) is a promising control signal for assist-as-needed hand rehabilitation after stroke, but detecting intent from paretic muscles often requires lengthy, subject-specific calibration and remains brittle to variability. We propose a healthy-to-stroke adaptation pipeline that initializes an intent detector from a model pretrained on large-scale able-bodied sEMG, then fine-tunes it for each stroke participant using only a small amount of subject-specific data. Using a newly collected dataset from three individuals with chronic stroke, we compare adaptation strategies (head-only tuning, parameter-efficient LoRA adapters, and full end-to-end fine-tuning) and evaluate on held-out test sets that include realistic distribution shifts such as within-session drift, posture changes, and armband repositioning. Across conditions, healthy-pretrained adaptation consistently improves stroke intent detection relative to both zero-shot transfer and stroke-only training under the same data budget; the best adaptation methods improve average transition accuracy from 0.42 to 0.61 and raw accuracy from 0.69 to 0.78. These results suggest that transferring a reusable healthy-domain EMG representation can reduce calibration burden while improving robustness for real-time post-stroke intent detection. Our project website, video, code, and dataset are available at: \url{https://roamlab.github.io/reactemg-stroke/}.
\end{abstract}

%%%%%%%%%%%%%%%%%%%%%%%%%%%%%%%%%%%%%%%%%%%%%%%%%%%%%%%%%%%%%%%%%%%%%%%%%%%%%%%%
%%%%%%%%%%%%%%%%%%%%%%%%%%%%%%%%%%%%%%%%%%%%%%%%%%%%%%%%%%%%%%%%%%%%%%%%%%%%%%%%
%%%%%%%%%%%%%%%%%%%%%%%%%%%%%%%%%%%%%%%%%%%%%%%%%%%%%%%%%%%%%%%%%%%%%%%%%%%%%%%%

\section{INTRODUCTION}
\label{sec:intro}

Robot-assisted wearable devices can provide task-specific, high-dose training to support post-stroke upper-limb recovery. In hand rehabilitation, these systems often assist finger extension (hand opening), allowing patients to repeatedly practice grasp-and-release movements~\cite{Park2020}. 

To better translate repetitions into recovery, assistance should engage the patient’s own neuromuscular drive rather than delivering purely passive motion~\cite{Hogan2006, Hu2009, Wolbrecht2008}. Many devices therefore perform intent detection, using biosignals to infer when a patient is attempting a movement. In this context, models that decode movement intent from surface electromyography (sEMG) are particularly attractive because sEMG is non-invasive, reflects muscle activation, and supports assist-as-needed control that builds on voluntary effort.

However, recent reviews emphasize that sEMG-based intent detection models for stroke survivors still rely on lengthy, subject-specific calibration and remain a key bottleneck for practical deployment~\cite{Gantenbein2022,Meyers2024}. Robust sEMG-based intent detection from paretic muscles typically requires such calibration because the underlying signals differ substantially from those of able-bodied users. Weakness and reduced voluntary drive produce small, noisy signals, while spasticity and abnormal synergies induce co-contraction (e.g., flexor activation during attempted hand opening)~\cite{Bandini2023}. As a result, each patient often requires large amounts of EMG data paired with ground-truth intent labels to obtain a usable model, and these models can be fragile. Performance may degrade with small changes in posture or sensor placement, and learned models risk overfitting to session-specific patterns that may not generalize reliably even to the same individual.

To address this bottleneck, we propose \emph{healthy-to-stroke few-shot adaptation} for sEMG intent detection: we initialize the intent detector from a model pretrained on large-scale able-bodied sEMG and then fine-tune it for each stroke survivor using only a small dataset. This reframes training as adapting a shared, reusable EMG representation to a new stroke participant, with most representation learning occurring in the healthy domain.

Most prior work addresses this bottleneck with few-shot or adaptive learning strategies in the stroke domain. Sarwat et al. apply a prototypical network that learns an embedding space and performs one-shot adaptation with distance-based classification~\cite{Sarwat2024}. La Rotta et al. propose MetaEMG, leveraging a meta-learning framework for rapid adaptation with a small subject-specific training set~\cite{LaRotta2024}. Xu et al. introduce ChatEMG, a generative model that produces synthetic stroke EMG samples to improve classifier performance~\cite{Xu2025}. 

In parallel, several studies leverage data sources beyond the stroke domain by constructing mixed healthy-stroke datasets. Cesqui et al. and Montecinos et al. perform a design-space search using healthy EMG, then apply the resulting healthy-derived configurations when training a model on stroke data~\cite{Cesqui2013,Montecinos2025}. Anastasiev et al. train a model on bilateral EMG from both the paretic and non-paretic limbs of stroke patients in a single stage, using the non-paretic limb as an additional subject-specific data source~\cite{Anastasiev2025}.

Taken together, these stroke-only and mixed-domain methods show that careful learning strategies can improve data efficiency. Yet they share a core limitation: the stroke model is still initialized from scratch and trained on small, stroke-only datasets. Even when healthy or non-paretic EMG is incorporated, it is typically used either to tune model configurations or as an auxiliary source for joint training, rather than to pretrain a reusable representation that can be transferred to new stroke users. Meanwhile, large sEMG datasets and foundation-style models for able-bodied users have emerged, demonstrating strong cross-subject performance and zero-shot generalization~\cite{ReactEMG2025,Kaifosh2025}. Since these models capture structures in EMG signals across many healthy users, they provide a natural source of transferable representations for the data-limited stroke setting. A direct way to exploit this structure is to initialize stroke intent detection from a healthy-pretrained model and adapt it via fine-tuning.

Motivated by this gap, we introduce ReactEMG Stroke, a healthy-to-stroke fine-tuning framework for sEMG intent detection. We treat ReactEMG, a model pretrained on able-bodied subjects, as an EMG foundation model and study its adaptation to stroke subjects in the context of intent detection for controlling a robotic hand orthosis. We ask how much per-subject labeled stroke data is needed when starting from a healthy model, and how different fine-tuning strategies behave under the same data budget. Our contributions are as follows:

\begin{itemize}
    \item \textbf{Healthy-to-stroke transfer learning for intent detection.}
    To the best of our knowledge, we are the first to adapt a healthy-domain EMG model to stroke intent detection. We fine-tune a pretrained healthy-domain model and show that it improves intent detection performance over stroke-only training across both in-domain and distribution-shifted test sets.
    \item \textbf{Comparison of fine-tuning strategies.}
    Under identical data budgets, training objectives, and model selection procedures, we compare stroke-only training from scratch against head-only fine-tuning, parameter-efficient low-rank adapters, and full-model fine-tuning. We evaluate accuracy and real-time control metrics across participants to distill practical guidance on how best to adapt a healthy model under tight budgets.
\end{itemize}

%%%%%%%%%%%%%%%%%%%%%%%%%%%%%%%%%%%%%%%%%%%%%%%%%%%%%%%%%%%%%%%%%%%%%%%%%%%%%%%%
%%%%%%%%%%%%%%%%%%%%%%%%%%%%%%%%%%%%%%%%%%%%%%%%%%%%%%%%%%%%%%%%%%%%%%%%%%%%%%%%
%%%%%%%%%%%%%%%%%%%%%%%%%%%%%%%%%%%%%%%%%%%%%%%%%%%%%%%%%%%%%%%%%%%%%%%%%%%%%%%%
\section{METHOD}
\label{sec:method}

Our goal is to infer a stroke participant’s intended hand gesture from forearm EMG signals recorded during attempted movement and use it to control a robotic hand orthosis that assists the corresponding gesture. We capture $8$-channel EMG signals (Myo armband, Thalmic Labs), sampled at $200$~Hz, and train a model to predict the current intent from a predefined set of $K$ action classes (e.g., \texttt{open hand}, \texttt{close hand}, \texttt{relax}) using short windows of muscle activity. Let $x=\{x_1, x_2, \dots, x_T\}\in\mathbb{R}^{T\times C}$ represent a window of length $T$ of multichannel EMG time series sampled from $C$ sensors. Given $x$, the model learns a mapping $f:\mathbb{R}^{T\times C}\rightarrow\mathbb{R}^{K}$ and outputs an intent likelihood vector $y\in\mathbb{R}^{K}$. In online use, $f$ is queried repeatedly on a sliding window over the incoming EMG stream to provide control signals for an orthosis.

We study healthy-to-stroke transfer learning for sEMG-based intent detection under data constraints. For each stroke participant, we start from a model pretrained on able-bodied data and fine-tune it using only a small, subject-specific training set. We then compare multiple adaptation strategies to a stroke-only baseline trained from scratch, and evaluate performance on held-out recordings under controlled distribution shifts: posture changes, armband repositioning, within-session drift, and orthosis actuation.

\subsection{Base Model}
\label{subsec:base_model}

Recent large-scale sEMG models have shown strong zero-shot generalization to unseen users in hand gesture recognition~\cite{ReactEMG2025, Kaifosh2025}. Among these approaches, we select ReactEMG as our healthy base model because it is designed specifically for real-time online use with a commercially available EMG armband~\cite{ReactEMG2025}. ReactEMG is an encoder-only transformer that treats both EMG signals and intent labels as input modalities. Using masked modeling, it learns to condition the EMG embeddings on the intent labels and jointly reconstructs both modalities as the output. The ReactEMG base model is pretrained on five public datasets that together provide labeled EMG recordings from more than 650 able-bodied participants.

\subsection{Stroke Dataset}
\label{subsec:dataset}

We collected a new sEMG dataset from three individuals with chronic stroke who attempted hand opening and closing with the paretic limb while wearing the MyHand orthosis~\cite{Park2020}. All participants provided informed consent to procedures approved by the Columbia University Institutional Review Board (Protocol \#AAAS8104). Participant information are summarized in \autoref{tab:participant_info}.

\begin{table}[h]
\centering
\caption{Participant information.}
\label{tab:participant_info}
\begin{tabular}{lccc}
\toprule
 & S1 & S2 & S3 \\
\midrule
Age                    & 56 & 58 & 74 \\
Gender                 & F  & F  & F  \\
Affected side          & R  & L  & L  \\
Years since stroke     & 6  & 4  & 10 \\
FMA--UE                & 26 & 35 & 34 \\
Hand subscore          & 1  & 2  & 8  \\
\bottomrule
\end{tabular}
\end{table}

\begin{figure}[t!]
\centering
\includegraphics[width=\columnwidth]{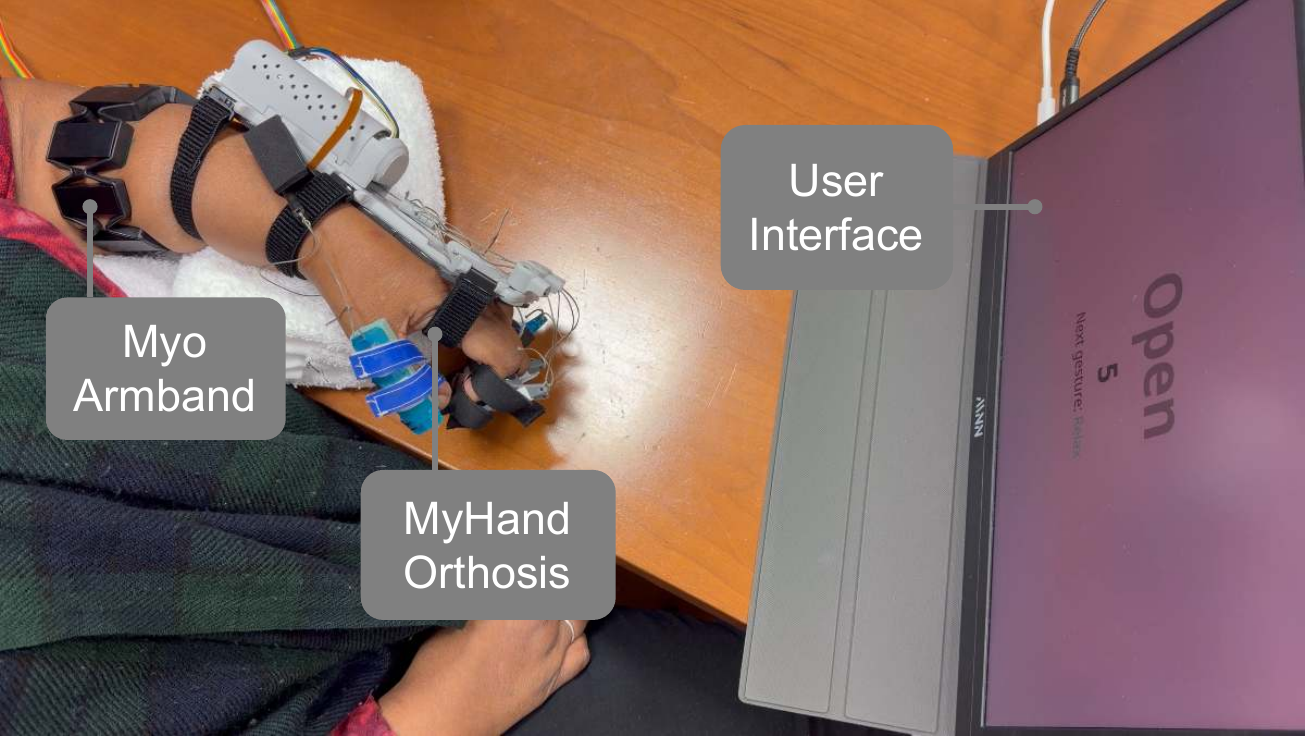}
\caption{\textbf{Stroke data collection setup.} a participant wears a Myo armband on the paretic forearm and the MyHand orthosis while an user interface presents timed movement cues and records synchronized labeled EMG.}
\label{fig:data}
\end{figure}

Each \emph{set} consists of two back-to-back timed instruction sequences: an opening sequence \texttt{ROROROR} followed by a closing sequence \texttt{RCRCRCR}, where \texttt{R} denotes relax, \texttt{O} attempted hand opening, and \texttt{C} attempted hand closing. Relax segments lasted 5~s, while \texttt{O}/\texttt{C} segments lasted 6~s to accommodate delayed recruitment and slower rises in effort commonly observed post-stroke~\cite{Seo2009}. Participants were instructed to initiate the attempt at cue onset and maintain a comfortable effort level until cue offset. \autoref{fig:data} shows the experimental setup.

Each participant completed nine \emph{sets} in total: four \emph{training sets} and five \emph{test sets}. The four training sets were collected in a standardized posture, with the forearm supported on a table and the participant's back against the chair backrest. The five test sets were designed to evaluate realistic sources of distribution shift:
\begin{itemize}
    \item \textbf{Within-session drift:} We repeated the standard configuration for two test sets---one mid-session and one at the end of the session---to capture signal drift due to fatigue and increased spasticity during the session.
    \item \textbf{Unseen posture:} We recorded one set with the arm lifted off the table, requiring the participant to maintain a hovering posture and support the arm's weight. This high-effort position changes the EMG distribution as additional muscles must be engaged.
    \item \textbf{Sensor placement:} We recorded one set with the Myo armband rotated approximately $15^\circ$ counterclockwise to assess robustness to imprecise sensor placement.
    \item \textbf{Device-driven motion:} We collected a closing sequence using the same \texttt{RCRCRCR} pattern with the MyHand orthosis powered. During each \texttt{R} (relax) segment immediately preceding a \texttt{C} cue, the MyHand orthosis opened the hand, and the participant attempted to close the hand from the open position at \texttt{C} onset.
\end{itemize}

\begin{table*}[t]
\centering
\small
\setlength{\tabcolsep}{4pt}
\renewcommand{\arraystretch}{1.15}
\caption{Intent detection performance for three stroke participants (S1--S3) averaged over five held-out test
sets.}
\label{tab:main_result}

\begin{tabular}{@{}l*{8}{c}@{}}
\toprule
& \multicolumn{2}{c}{\textbf{S1}} & \multicolumn{2}{c}{\textbf{S2}} & \multicolumn{2}{c}{\textbf{S3}} &
\multicolumn{2}{c}{\textbf{Avg.}} \\
\cmidrule(lr){2-3}\cmidrule(lr){4-5}\cmidrule(lr){6-7}\cmidrule(lr){8-9}
\textbf{Method}
& \textbf{Raw Acc.}   & \textbf{Trans. Acc.}
& \textbf{Raw Acc.}   & \textbf{Trans. Acc.}
& \textbf{Raw Acc.}   & \textbf{Trans. Acc.}
& \textbf{Raw Acc.}   & \textbf{Trans. Acc.} \\
\midrule
Zero-shot   & $0.60 \pm 0.02$ & $0.05 \pm 0.07$ & $0.56 \pm 0.08$ & $0.22 \pm 0.19$ & $0.63 \pm 0.06$ & $0.13 \pm
0.13$ & $0.60$ & $0.13$ \\
\midrule
Stroke-only & $\boldsymbol{0.71 \pm 0.05}$ & $0.28 \pm 0.14$ & $0.61 \pm 0.20$ & $0.32 \pm 0.20$ & $0.74 \pm
0.13$ & $0.67 \pm 0.28$ & $0.69$ & $0.42$ \\
\midrule
Head-only   & $0.65 \pm 0.20$ & $0.33 \pm 0.41$ & $0.71 \pm 0.06$ & $0.43 \pm 0.25$ & $\boldsymbol{0.89 \pm
0.07}$ & $\boldsymbol{0.83 \pm 0.20}$ & $0.75$ & $0.53$ \\
LoRA        & $0.70 \pm 0.24$ & $\boldsymbol{0.45 \pm 0.28}$ & $\boldsymbol{0.78 \pm 0.08}$ & $\boldsymbol{0.62
\pm 0.22}$ & $0.88 \pm 0.08$ & $0.75 \pm 0.21$ & $\mathbf{0.78}$ & $\mathbf{0.61}$ \\
Full        & $\boldsymbol{0.71 \pm 0.16}$ & $0.40 \pm 0.16$ & $0.75 \pm 0.14$ & $\boldsymbol{0.62 \pm 0.25}$ &
$0.87 \pm 0.08$ & $0.82 \pm 0.21$ & $\mathbf{0.78}$ & $\mathbf{0.61}$ \\
\bottomrule
\end{tabular}
\end{table*}

\subsection{Fine-tuning Strategies and Baselines}
\label{subsec:finetune}

ReactEMG~\cite{ReactEMG2025} is an encoder-only transformer pretrained with a multimodal masked-modeling objective in which the EMG stream and the aligned intent stream are embedded, jointly processed by the transformer, and partially masked during training. The model learns to reconstruct the masked portions, encouraging representations that align muscle activity with intended actions. After the encoder, linear heads produce per-timestep intent logits and an auxiliary EMG reconstruction output used for pretraining.

For our fine-tuning experiments, we used the same preprocessing, training objective, and input/output structure as ReactEMG. We refer to the EMG embedding and the transformer encoder stack as the \emph{backbone}, and the final linear layer that maps encoder features to $K$ intent logits as the \emph{classification head}. We evaluated three fine-tuning strategies for adapting the healthy-pretrained ReactEMG model to stroke: head-only fine-tuning, LoRA, and full fine-tuning.

\subsubsection{Head-only Fine-tuning}
For head-only fine-tuning, we froze the pretrained ReactEMG backbone and updated only the classification head, initializing the head from pretrained weights. Since this approach modifies only a small set of parameters, it is fast and less prone to overfitting, but it may underperform when stroke EMG deviates substantially from healthy patterns.

\subsubsection{LoRA}
In LoRA fine-tuning, we froze all pretrained weights and trained low-rank update matrices for all linear layers in the network. Adaptation is achieved through a small number of additional parameters rather than by modifying the full weight tensors~\cite{Hu2021}. This provides greater capacity to adjust internal features to paretic EMG patterns while keeping trainable parameters and storage small. However, low-rank updates could still be too restrictive when the healthy-to-stroke shift requires larger representation changes. 

\subsubsection{Full Fine-tuning}
In full-model fine-tuning, we updated all model parameters. This offers the greatest flexibility to match stroke-specific EMG distributions, but is also the most compute-intensive and the most susceptible to overfitting and forgetting in the few-shot regime. 

We compare fine-tuning strategies against two baselines: 
\subsubsection{Healthy Zero-shot}
The healthy model is the frozen pretrained ReactEMG model evaluated zero-shot on stroke.

\subsubsection{Stroke-only Training}
We train a stroke-only ReactEMG baseline from scratch (random weight initialization) using the same training set as the fine-tuning approaches.

These setups are illustrated in \autoref{fig:intro}. We refer readers to our codebase for a complete list of parameters used for preprocessing, training, and evaluation.

%%%%%%%%%%%%%%%%%%%%%%%%%%%%%%%%%%%%%%%%%%%%%%%%%%%%%%%%%%%%%%%%%%%%%%%%%%%%%%%%
%%%%%%%%%%%%%%%%%%%%%%%%%%%%%%%%%%%%%%%%%%%%%%%%%%%%%%%%%%%%%%%%%%%%%%%%%%%%%%%%
%%%%%%%%%%%%%%%%%%%%%%%%%%%%%%%%%%%%%%%%%%%%%%%%%%%%%%%%%%%%%%%%%%%%%%%%%%%%%%%%

\section{EVALUATION}
\label{sec:evaluation}

\subsection{Experimental Setup}
\label{subsec:exp_setup}

For all experiments, we train a separate model per stroke participant using only that participant’s training data, and evaluate it on held-out data from the same participant under distribution shift. For each trainable model variant, we select hyperparameters (e.g., learning rate, regularization strength, and number of epochs) using grid search with 4-fold cross-validation over the four training sets. We compute accuracy on the held-out set for each fold and average these accuracies across folds for each hyperparameter configuration. We select the configuration with the highest validation accuracy, retrain once on all four training sets with the selected hyperparameters, and evaluate on the five held-out test sets. All training and evaluation configurations are available in our codebase.

\subsubsection{Data Efficiency}
To study data efficiency in the healthy-to-stroke setting, we fix the adaptation method and hyperparameters to the best-performing fine-tuning strategy found by cross-validation, and vary how much stroke training data is used for fine-tuning. Each subject's training dataset is split into 12 pairs, with each pair containing one open attempt and one close attempt. For a given data budget $N \in \{1,4,8\}$, we sample $N$ pairs without replacement from the 12 available pairs, train a model using only those $N$ pairs, and evaluate on the same held-out test sets as before. We repeat the sampling process 12 times for each data budget and record the average accuracy over the samplings.

\subsubsection{Convergence Behavior}
To study convergence behavior, we extend training beyond the cross-validated schedule. For each trainable model variant and participant, we take the selected hyperparameter configuration and train for $100$ epochs under the same optimization settings. We save a checkpoint once every 5 epochs and evaluate each checkpoint on the stroke participant’s test sets. We also evaluate the frozen healthy-pretrained model (no adaptation) on the same stroke test sets as a reference. This produces accuracy curves that show how stroke performance evolves with additional fine-tuning and when diminishing returns occurs.

\subsection{Metrics}
We evaluate our model in the same way as ReactEMG to capture both standard accuracy measures and measures that better reflect real-time use:
\begin{itemize}
    \item \textbf{Raw accuracy}: fraction of timesteps whose predicted label matches the cue-derived intent label.
    \item \textbf{Transition accuracy}: fraction of ground-truth intent transitions that are correct under the ReactEMG definition~\cite{ReactEMG2025}, requiring at least one correct prediction of the new class within a short reaction buffer around the transition and error-free maintenance of that class throughout the subsequent maintenance period until the next transition.
\end{itemize}

We report transition accuracy because it is designed to reflect two failure modes that affect online real-time control: (i) delayed switching when the user changes intent, and (ii) brief ``flicker'' during sustained intent that can cause unstable device behavior. Raw accuracy can obscure these issues because both delayed onsets and flickers contribute to only a small number of per-frame errors, and long \texttt{relax} periods can dominate the data and inflate accuracy even when \texttt{open}/\texttt{close} commands are unreliable. Transition accuracy instead treats each ground-truth intent change as an event, allowing a short reaction buffer around the transition boundary and then requiring stable, error-free maintenance of the new command until the next transition. This results in a strict but more control-relevant measure of whether a model feels stable and responsive in practice.

%%%%%%%%%%%%%%%%%%%%%%%%%%%%%%%%%%%%%%%%%%%%%%%%%%%%%%%%%%%%%%%%%%%%%%%%%%%%%%%%
%%%%%%%%%%%%%%%%%%%%%%%%%%%%%%%%%%%%%%%%%%%%%%%%%%%%%%%%%%%%%%%%%%%%%%%%%%%%%%%%
%%%%%%%%%%%%%%%%%%%%%%%%%%%%%%%%%%%%%%%%%%%%%%%%%%%%%%%%%%%%%%%%%%%%%%%%%%%%%%%%

\begin{figure*}[t]
\centering
\includegraphics[width=0.8\textwidth]{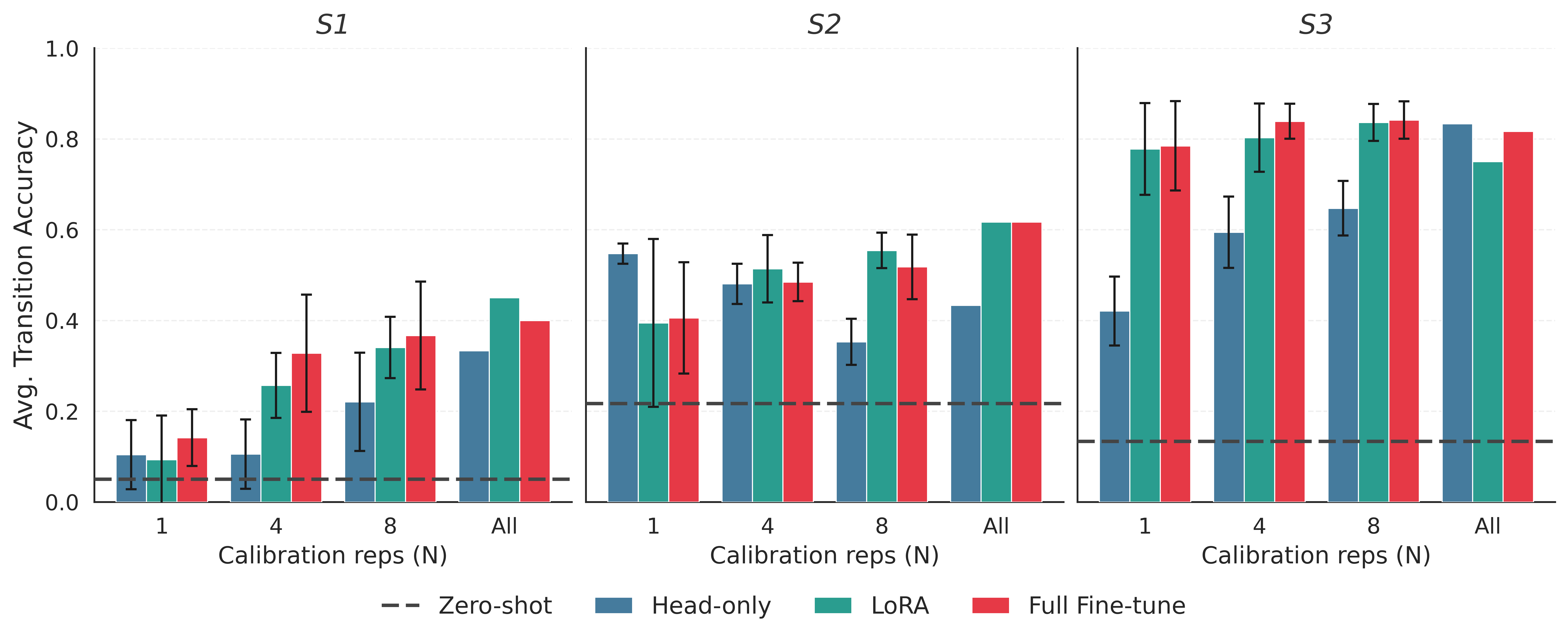}
\caption{\textbf{Data Efficiency Comparison}. Average transition accuracy across different data budgets for all three fine-tuning methods. The dotted zero-shot line indicates frozen healthy base model evaluated on stroke test sets. Error bars represent standard deviation over repeated samplings.}
\label{fig:data_efficiency}
\end{figure*}

\section{RESULTS AND DISCUSSION}
\label{sec:results}

\subsection{Healthy pretraining improves intent detection under realistic distribution shift}
\label{subsec:overall_performance}
Table~\ref{tab:main_result} summarizes intent detection performance averaged over five held-out test sets. The frozen healthy-pretrained model exhibits non-trivial zero-shot transfer to stroke, but its transition accuracy is very low, illustrating that per-timestep correctness does not necessarily translate to reliable real-time control. Training a stroke-only model from scratch improves performance, confirming the need for subject-specific training data. However, initializing from the healthy-pretrained ReactEMG model and adapting to each stroke subject yields further gains. Across all three participants, healthy-initialized adaptation improves transition accuracy over stroke-only training, and the best-performing adaptation methods (LoRA and full fine-tuning) achieve an improvement of $+0.09$ on raw accuracy and $+0.19$ on transition accuracy.

\subsection{Fine-tuning strategy depends on the participant}
\label{subsec:finetune_comparison}
Head-only fine-tuning (frozen backbone, trainable classifier) improves average transition accuracy to $0.53$, indicating that the pretrained backbone provides a useful representation for stroke intent detection. However, methods that can more aggressively adapt internal representations produce the strongest overall performance: LoRA and full fine-tuning tie on the across-subject average. This observation also suggests that parameter-efficient low-rank updates can capture much of the healthy-to-stroke shift without updating all model parameters, which is appealing for training under limited compute and storage budgets.

The differences in accuracy across subjects aligns qualitatively with impairment differences in \autoref{tab:participant_info}: absolute performance is lowest for the more impaired participant S1 and highest for the less impaired participant S3. At the same time, there is no single fine-tuning strategy that dominates for every participant. For S1 and S2, the best transition accuracy requires LoRA / Full Fine-tuning, whereas for S3 the highest transition accuracy is achieved by head-only tuning. 

One plausible hypothesis is that the best fine-tuning strategy depends on the level of impairment of the participant: participants with less severe hand impairment may be closer to the healthy EMG manifold and therefore require less aggressive adaptation, while participants with more severe impairment may need deeper representation changes. We emphasize, however, that this remains a hypothesis, and a larger cohort is needed to draw confident conclusions.

\subsection{Data efficiency}
\label{subsec:data_efficiency}
To evaluate data efficiency, we vary the labeled stroke data budget $N \in \{1,4,8\}$ and compare the resulting models against the frozen healthy baseline ($N=0$, zero-shot) and training with all available training data ($N=\mathrm{All}$), as shown in \autoref{fig:data_efficiency}. Across all participants, healthy-to-stroke adaptation produces improvements even at very low data budgets. The improvements are mostly front-loaded with diminishing returns as $N$ increases, especially for LoRA and Full Fine-tuning. The magnitude of the single-sample ($N=1$) improvement is participant-dependent: for the less impaired participants S2 and S3, transition accuracy increases sharply and approaches (sometimes surpasses) the performance of full-budget training. In contrast, S1's model improves more gradually and continue to benefit from larger budgets. Overall, these results indicate that healthy-pretrained adaptation can substantially reduce data burden for some individuals, but that the amount of labeled data required remains heterogeneous across stroke participants.

\begin{figure}[t]
\centering
\includegraphics[width=\linewidth]{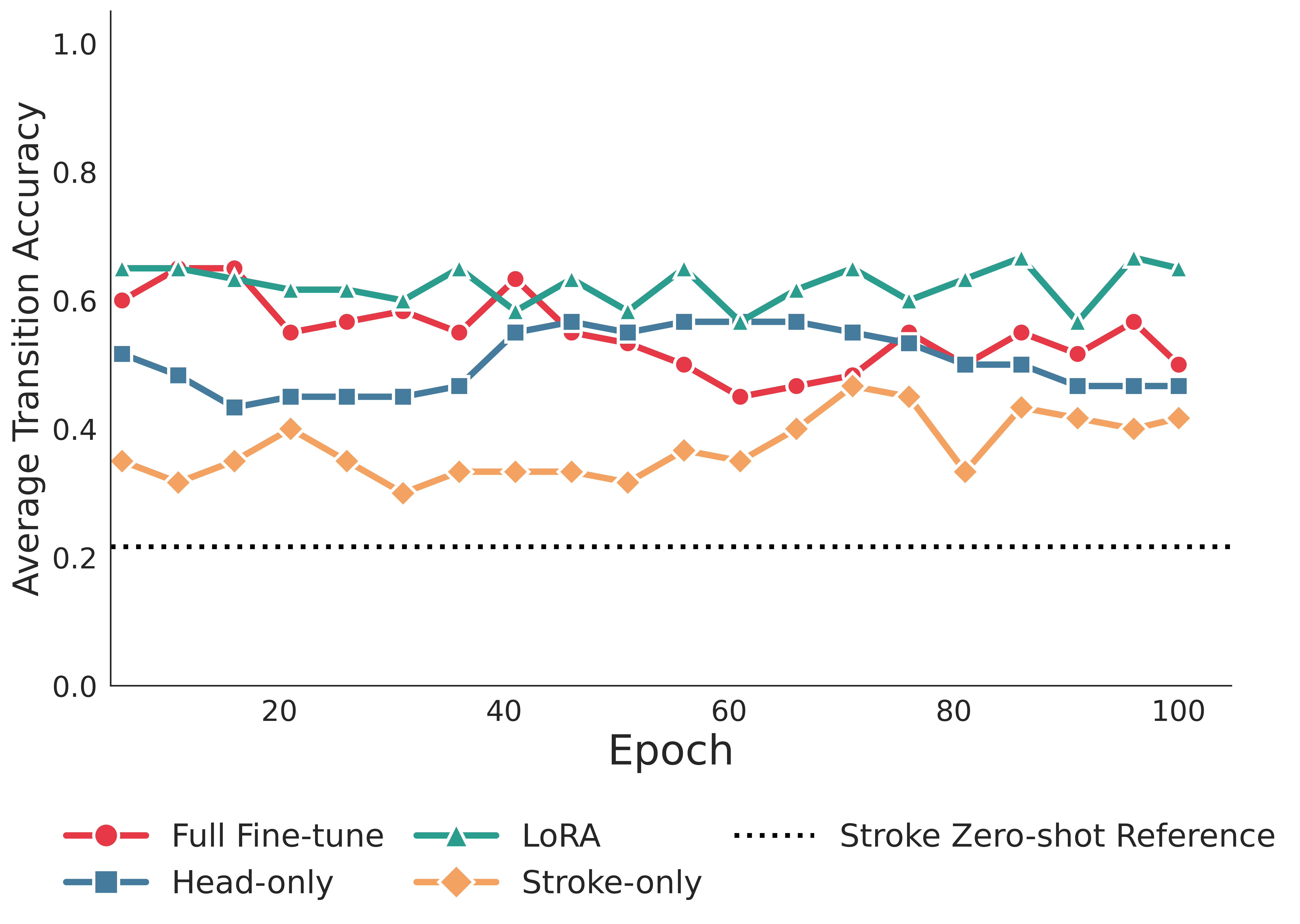}
\caption{\textbf{Convergence behavior on S2}. Average transition accuracy on the five held-out S2 test sets versus training epoch for stroke-only training and three healthy-initialized adaptation strategies. Dotted lines denote zero-shot performance of the frozen healthy-pretrained model on stroke.}
\label{fig:convergence}
\end{figure}

\subsection{Convergence behavior}
\autoref{fig:convergence} shows accuracy curves for participant S2 when each trainable variant is trained for 100 epochs using the cross-validated hyperparameters and evaluated every 5 epochs. All healthy-initialized adaptation methods exceed the stroke zero-shot reference within the first checkpoint, indicating that only a small number of fine-tuning iterations is needed to obtain meaningful gains in transition accuracy. In contrast, the stroke-only model trained from scratch improves more gradually, suggesting that healthy pretraining provides a better starting representation under the same limited data and compute budgets.

\section{CONCLUSION}
\label{sec:conclusion}
We presented \emph{ReactEMG Stroke}, a healthy-to-stroke few-shot adaptation framework for sEMG-based intent detection to support assist-as-needed hand rehabilitation after stroke. Our results suggest that initializing from a healthy-pretrained model and fine-tuning with only a small amount of subject-specific stroke data can outperform both zero-shot transfer and stroke-only training, even under distribution shifts. Beyond these performance gains, healthy-to-stroke fine-tuning is also appealing from a scaling perspective: although it is not guaranteed that scaling a healthy foundation model will help stroke fine-tuning more than scaling a stroke-only model, any benefit from model scaling is much easier to realize in the healthy domain, where EMG can be collected safely and cheaply at scale. In contrast, stroke EMG will remain inherently scarce, making healthy-to-stroke fine-tuning an attractive strategy for translating large model capacity into better intent detection from limited stroke data.

%%%%%%%%%%%%%%%%%%%%%%%%%%%%%%%%%%%%%%%%%%%%%%%%%%%%%%%%%%%%%%%%%%%%%%%%%%%%%%%%

\bibliographystyle{IEEEtran}
\bibliography{reactemg_biorob}

\end{document}